\documentclass[11pt]{article}

\usepackage{acl}

\usepackage{times}
\usepackage{latexsym}
\usepackage{changes}
\usepackage[T1]{fontenc}

\usepackage[utf8]{inputenc}

\usepackage{microtype}
\usepackage{enumitem}
\usepackage{inconsolata}

\usepackage{graphicx} 
\usepackage{newfloat}
\usepackage{listings}
\usepackage{array}
\usepackage{multirow}
\usepackage{booktabs}
\usepackage{amssymb}

\newcommand{\vpara}[1]{\vspace{0.05in}\noindent\textbf{#1 }}
\newcommand{\hide}[1]{} 

\newcommand{\emojisob}{$\vcenter{\hbox{\includegraphics[width=2ex,height=2ex]{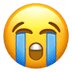}}}$}
\newcommand{\emojicar}{$\vcenter{\hbox{\includegraphics[width=2ex,height=2ex]{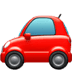}}}$}

\newcommand{\emojiheart}{$\vcenter{\hbox{\includegraphics[width=2ex,height=2ex]{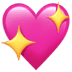}}}$}

\newcommand{\emojiptwx}{$\vcenter{\hbox{\includegraphics[width=2ex,height=2ex]{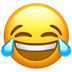}}}$}

\newcommand{\emojidancers}
{$\vcenter{\hbox{\includegraphics[width=2ex,height=2ex]{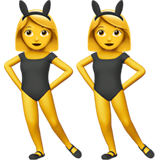}}}$}
\newcommand{\emojiwatermelon}{$\vcenter{\hbox{\includegraphics[width=2ex,height=2ex]{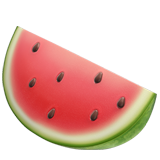}}}$}

\newcommand{\emojicrown}{$\vcenter{\hbox{\includegraphics[width=2ex,height=2ex]{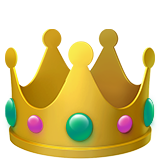}}}$}
\newcommand{\emojiseenoevil}{$\vcenter{\hbox{\includegraphics[width=2ex,height=2ex]{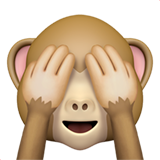}}}$}
\newcommand{\emojiconfused}{$\vcenter{\hbox{\includegraphics[width=2ex,height=2ex]{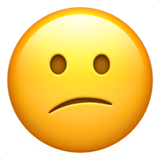}}}$}

\title{Unleashing the Power of Emojis in Texts via Self-supervised Graph
Pre-Training}

\author{
 \textbf{Zhou Zhang\textsuperscript{1}}\footnotemark[1],
 \textbf{Dongzeng Tan\textsuperscript{2}}\footnotemark[1],
 \textbf{Jiaan Wang\textsuperscript{1}},
 \textbf{Yilong Chen\textsuperscript{1}},
 \textbf{Jiarong Xu\textsuperscript{1}}\footnotemark[2]
\\
\\
 \textsuperscript{1}Fudan University,
 \textsuperscript{2}Alibaba Group
\\
 \small{
   \textbf{Correspondence:} 
   \href{mailto:jiarongxu@fudan.edu.cn}{zhouzhang23@m.fudan.edu.cn}, \href{mailto:erictandz@gmail.com}{erictandz@gmail.com}, \href{mailto:jawang.nlp@gmail.com}{jawang.nlp@gmail.com},
 }
 \\
 \small{
   \href{mailto:yilongchen@uchicago.edu}{yilongchen@uchicago.edu}, 
   \href{mailto:jiarongxu@fudan.edu.cn}{jiarongxu@fudan.edu.cn}
 }
}

\begin{document}
\maketitle

\renewcommand{\thefootnote}{\fnsymbol{footnote}}
\footnotetext[1]{Both authors contributed equally to this research.}
\footnotetext[2]{Corresponding author.}

\renewcommand{\thefootnote}{\arabic{footnote}}

\begin{abstract}
Emojis have gained immense popularity on social platforms, serving as a common means to supplement or replace text.
However, existing data mining approaches generally either completely ignore or simply treat emojis as ordinary Unicode characters, which may limit the model's ability to grasp the rich semantic information in emojis and the interaction between emojis and texts. 
Thus, it is necessary to release the power of emojis in social media data mining.
To this end, we first construct a heterogeneous graph consisting of three types of nodes, \emph{i.e.}, post, word and emoji nodes to improve the representation of different elements in posts. The edges are also well-defined to model how these three elements interact with each other. To facilitate the sharing of information among post, word and emoji nodes, we propose a graph pre-training framework for text and emoji co-modeling, which contains two graph pre-training tasks: node-level graph contrastive learning and edge-level link reconstruction learning. 
%
Extensive experiments on the Xiaohongshu and Twitter datasets with two types of downstream tasks demonstrate that our approach proves significant improvement over previous strong baseline methods.\footnote{\href{https://github.com/ginkoeric/Self-supervised-Graph-Pre-training-for-Emoji}{https://github.com/ginkoeric/Self-supervised-Graph-Pre-training-for-Emoji}}
\end{abstract}

\section{Introduction}
\label{sec:intro}


Emojis have gained significant attention due to their popularity in digital communication, especially on social media platforms. They convey emotions and reactions, supplementing or replacing text in posts, which adds nuance to conversations. For example, the heart emoji \emojiheart $ $ conveys love more strongly than words, and the face-with-tears-of-joy emoji \emojiptwx $ $ effectively expresses humor. Accurately modeling emojis can enhance natural language processing (NLP) tasks like sentiment analysis and emoji generation.



Current methods for processing emojis are varied. Traditional statistical methods either ignore emojis or treat them as regular characters, failing to capture their semantics. Methods like emoji2vec \citep{Eisner2016emoji2vecLE} learn emoji representations based on textual descriptions but rely heavily on annotations, which fails to keep up with the evolving semantics of emojis in different contexts. Self-supervised learning, which doesn't require labeled data, is widely used for emojis. However, many self-supervised methods either use emojis only as annotations ~\citep{deepmoji} for text embeddings or focus solely on emoji embeddings, neglecting text embeddings. These models fail to fully utilize the relationship between text and emojis.

To address these issues, we aim to better model the relationship between emojis and text, improving embeddings by learning their interactions. It can be observed that emojis in similar texts share semantics, and texts with similar emojis often have shared sentiments or popularity.

Given the ability of heterogeneous graphs to model interactive relations across information at different granularities, we construct a heterogeneous graph to represent the information that lies in social media posts.
Specifically, the graph contains emoji nodes and word nodes to capture the relationship between emojis and texts. To better absorb the global information of the whole post, another node type, \emph{i.e.}, post nodes, is also constructed in our graph.
In this manner, the above similarity could be represented by the similarity between $n$-hop neighbors.


%
%
%
In order to make information further flow between different nodes in the heterogeneous graph, we develop two self-supervised graph pre-training tasks that facilitate reciprocal text-emoji joint representation learning.
%
%
(1) Based on the observations above, we construct \emph{node-level subgraph contrastive learning task}. This task first randomly chooses a starting node, and then samples its neighbors through random walk, subsequently forming a sub-graph containing both the starting node and the neighbors.
By repeating this process on different starting nodes, several sub-graphs are obtained. The task is used to guide the graph model to judge whether two sub-graphs are generated from the same starting node.
%
This task could help capture the similarity among nodes.
%
%
(2) To better capture the mutual relationships among these three types of nodes, we build the second task: \emph{edge-level link reconstruction learning}. This task selects existing edges as positive edges and constructs negative edges from randomly added non-existing ones. Then, the task requires the graph model to distinguish positive edges from negative ones.
%
This task lets the graph model focus on the relationships among different node types (every edge links two nodes from different types).
Consequently, with the help of the above two tasks, graph models could learn to understand how posts, emojis and words interact with each other, which will benefit downstream tasks like popularity prediction.

Our main contributions are concluded as follows:
\begin{itemize}[leftmargin=*,topsep=0pt]
\setlength{\itemsep}{0pt}
\setlength{\parsep}{0pt}
\setlength{\parskip}{0pt}
\item To better describe the relationship between posts, emojis and words in social media data, we construct a post-emoji-word heterogeneous graph.
\item We design a graph pre-training framework to learn the post, emoji, and word-level embedding. This framework consists of two self-supervised tasks on both nodes and edges in the heterogeneous graph. The embedding learned has strong transferability, which could be integrated with existing language models' text/token embedding, and improve their emoji understanding ability, and then be used in various downstream tasks.
\item We conduct extensive experiments on datasets from different social media platforms (\emph{i.e.}, Xiaohongshu and Twitter) to demonstrate that our approach achieves better performance on various downstream tasks (\emph{e.g.} popularity prediction and sentiment prediction) than strong baselines. Our model outperforms the baseline model by 2\% - 10\%. 

\end{itemize}




\section{Related Work}
\label{sec:related}
Emojis play multiple roles in people’s communication. They not only provide an expressive way to convey emotions or situational information but are also convenient for grasping implicit sentiments as well as emotions.
The use of emojis on social media tends to affect the popularity of tweets. However, the importance of emojis in textual studies has not been fully realized by early work~\citep{hotho2005brief,mikolov2013efficient,allahyari2017brief}.
Recently, many studies have revealed the significance of emojis, and these studies could be summarized into four types of research fields: traditional pre-process, description mining, supervised learning, and self-supervised learning methods.

\noindent \textbf{Traditional Pre-process Methods.} These methods directly utilize standard pre-processing methods and then feed the pre-processed emojis to language models~\citep{Wijeratne2017,pavan2019novel,Yinxia2020}. Specifically, 
(1) \emph{Removing emoji strategy} treats emojis as noise and remove them from the original text~\citep{pavan2019novel}. This strategy neglects the semantic information conveyed by emojis.
(2) \emph{Keeping emoji strategy} retains emojis as tokens, but unfortunately, the tokens fail to convey the accurate meanings carried by the emojis~\citep{Yinxia2020}.
(3) \emph{Translating emoji} translates emojis into descriptive texts~\citep{Wijeratne2017} and learn representations from these descriptions.
Though these methods are intuitive, to perform pre-processing on emojis, we have to maintain a dictionary that covers specific types of emojis (\emph{e.g.}, iOS emojis encoded by Unicode). Such a dictionary cannot be generalized to other types of emojis (\emph{e.g.}, Xiaohongshu emojis and user-created emojis). 


\noindent \textbf{Description Mining Methods.} To further capture the emojis' semantics, another line of research investigates methods for learning emoji representations by utilizing the corresponding textual descriptions~\citep{Eisner2016emoji2vecLE}. \citet{Yinxia2020} introduce an attention model to learn representations for emojis and text, while simultaneously conducting sentiment classification. This approach learns emoji representations by minimizing the similarity between the representation of an emoji and its corresponding textual description. However, the effectiveness of this approach relies on the availability of textual descriptions for emojis, which may not be accessible for certain types of emojis, such as user-created emojis. 

\noindent \textbf{Supervised Learning Methods.} The third line of work explores the use of emojis to assist in various downstream tasks, such as sentiment classification, emotion analysis, and emoji prediction~\citep{Chen2019,singh-etal-2022-emoji,zhao2023pedm}. Researchers create annotated datasets where text inputs are paired with corresponding emojis, enabling models to discern the connection between textual contexts and emojis implicitly. For instance, \citet{Chen2019} tackle the challenge of cross-lingual sentiment classification by using emojis as an instrument because similar emojis convey similar emotions across different languages.
These methods are limited to only leveraging the emoji meaning related to certain tasks at hand, \emph{e.g.}, emojis employed for sentiment classification may not effectively represent the semantic and concrete meaning of an emoji like ``\emojicar''.

\noindent \textbf{Self-Supervised Learning Methods.} The fourth line of existing studies focuses on utilizing emoji information and learning emoji representations via self-supervised learning. In detail, these methods leverage self-supervised (or unsupervised) tasks to incorporate emoji information into text representations and enhance the meaningfulness of emojis. For instance, \citet{deepmoji,park-etal-2018-plusemo2vec} propose DeepMoji which involves identifying whether a post contains specific emojis. \citet{10.1145/3338906.3338977} further develop SEntiMoji for sentiment analysis based on DeepMoji. These approaches allow for the integration of emoji information into text representations. Another study by \citet{emojicoocurrence} constructs an emoji co-occurrence network and learns emoji representations by minimizing the similarity between neighboring emojis.
Nevertheless, these methods have a limitation as they do not consider the positional relationship between the text and emojis. Also, emojis are used only as annotations in these methods and they are trained solely based on specific tasks. The lack of universal emoji embedding will make it difficult for these methods to be applied to a wide range of downstream tasks. To bridge this gap, our study investigates how to incorporate the positional information and utilize the relation between text and emojis to learn universal embeddings of emojis.

Different from previous studies, in this work, a heterogeneous graph is constructed to depict the connections among posts, words, and emojis. Besides, two self-supervised learning tasks are designed: node-level subgraph contrastive learning and edge-level link reconstruction learning, which facilitate the simultaneous refinement of posts, words, and emoji representations.
\section{Methodology}
\label{sec:methodology}


In this section, we first construct the heterogeneous graph with three types of nodes: posts, emojis and words, to capture the information and interactions among them (\S~\ref{subsec:3.1}). Then, based on the heterogeneous graph, we design a graph pre-training framework to model the comprehensive interaction among all types of nodes (\S~\ref{subsec:3.2}). After that, we show how to use our proposed method in various downstream tasks, \emph{i.e.}, popularity prediction 
and sentiment classification(\S~\ref{subsec:downstreamtask}). Figure~\ref{fig:graph} illustrates the overview of our pre-training and fine-tuning stages.

\subsection{Heterogeneous Graph Construction}
\label{subsec:3.1}
\begin{figure*}[t]
  \centering
  \includegraphics[width=\textwidth]{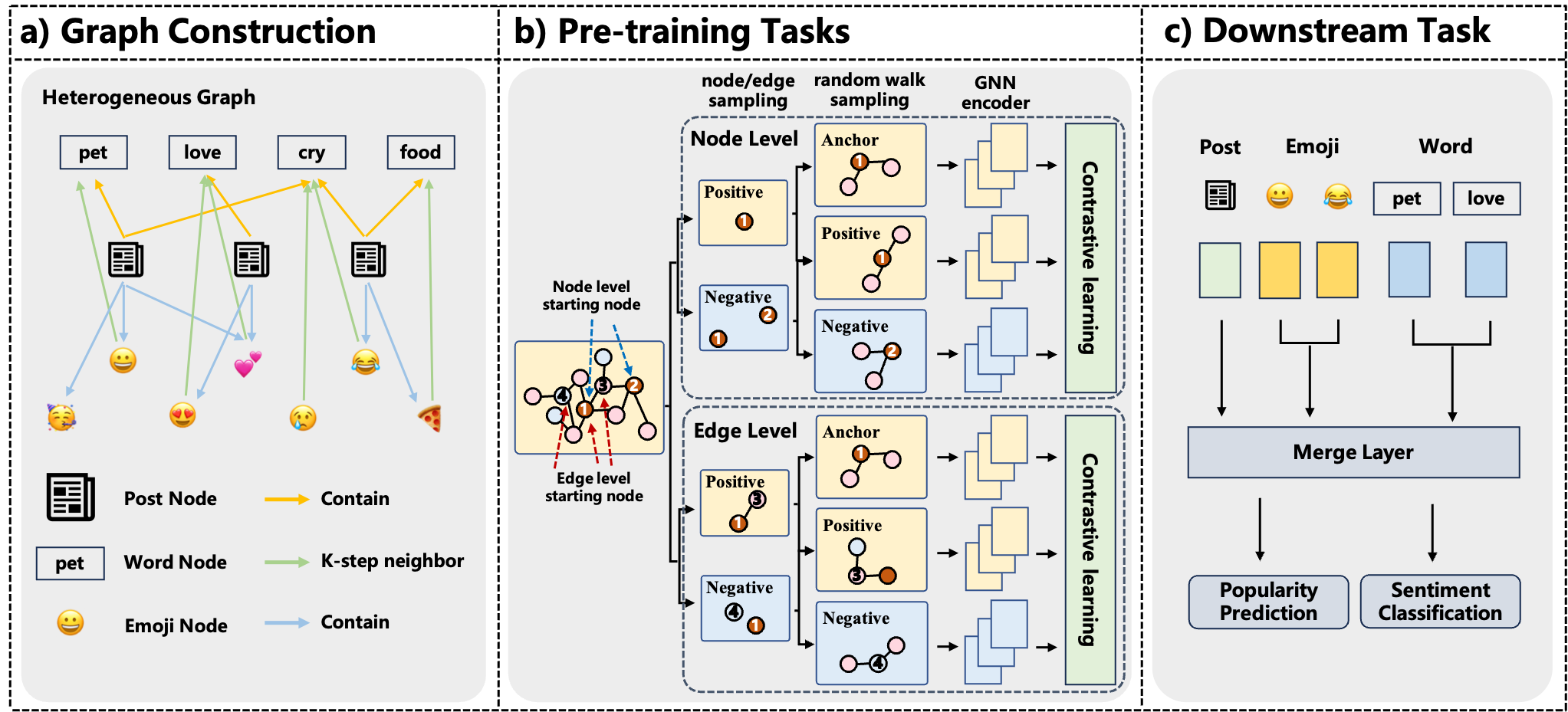}
  \caption{An illustration of all processes of our model. Part a) shows the construction of our heterogeneous graph that incorporates post, emoji and word nodes, along with their connections. Part b) shows the structure of node level and edge level pre-training tasks. Part c) shows the structure of two downstream tasks.}
  \label{fig:graph}
\end{figure*}

To incorporate the relationship and information between textual content and emojis, we build up a heterogeneous graph, as shown in Figure~\ref{fig:graph}(a), which consists of three types of nodes:
\begin{itemize}[leftmargin=*,topsep=0pt]
\setlength{\itemsep}{0pt}
\setlength{\parsep}{0pt}
\setlength{\parskip}{0pt}
\item \textbf{Post node}: $\mathcal{T}=\{ t_1,t_2,\cdots,t_{N}\}$ represents the set of post nodes, where $t_i (i=1,2,\cdots, N)$ is the $i$-th post node containing only the textual information in the $i$-th post, and $N$ is the total number of posts.
\item \textbf{Word node}: $\mathcal{W}=\{ w_1,w_2,\cdots,w_{M}\}$ is the set of word nodes (\emph{i.e.}, the word vocabulary), where $M$ is the vocabulary size. $\mathcal{W}$ includes all the words collected from all the posts. 
\item \textbf{Emoji node}: $\mathcal{E}=\{ e_1,e_2,\cdots,e_{K}\}$ means the set of emoji nodes, comprising the top $K$ emojis with the highest occurrences in posts.
\end{itemize}

Among these nodes, there exist three kinds of edges to describe their connections.
\begin{itemize}[leftmargin=*,topsep=0pt]
\setlength{\itemsep}{0pt}
\setlength{\parsep}{0pt}
\setlength{\parskip}{0pt}
\item \textbf{Post-emoji edge}: This type of edge indicates that if an emoji is in/out of the post. In this manner, we can build connections among posts that share the same emojis, and lead to the information flow from post to emoji. For example, posts ``I didn't pass the exam \emojisob'' and ``I finally pass the exam \emojisob'' tell totally different emotions with the same emoji \emojisob. The former is a sad feeling while the latter means an overwhelming joyful feeling opposite. Therefore, connections between posts and emojis will help a lot in an informative representation study for both.
\item \textbf{Post-word edge}: This edge type shows the existence of a word in the post. With no doubt that building connections between all words from the post and itself will make the graph complex and may introduce uncontrollable noise. To avoid that, we treat the words inside hashtag as keywords, and then filter keywords from posts and only consider links between them. Typically, the hashtag of a post matches its topic. In this way, we can make sure that connected posts will share the same topic, and make the textual semantic information more finely. 
\item \textbf{Word-emoji edge}: This type of edge comes from the emoji and its $k$-step neighbor words. For instance, in the sentence ``What a nice \emojiheart $ $ day!'', words ``a'', ``nice'', and ``day'' are \emojiheart's two-step neighbors. They contribute to a fine-grained semantic knowledge for emojis, and help us to locate emojis' position in a post.
\end{itemize}


To initialize the node embeddings, we use the pre-trained language model (i.e., BERT~\citep{devlin2018bert}) to generate representations for post nodes and word nodes. For emoji nodes, we obtain the initialization by randomization, these will be discussed comprehensively in Section~\ref{sec:experiment}.

\subsection{Graph Pre-training Framework}
\label{subsec:3.2}
Based on the heterogeneous graph built above, we design a graph pre-training framework to learn the post, emoji, and word-level representations.
The framework consists of two self-supervised tasks: node-level sub-graph contrastive learning and edge-level link reconstruction learning. Both of them are designed to enhance the ability to encode comprehensive semantics among posts, emojis, and words by the general GNN encoder model.

\vpara{Node-level Sub-graph Contrastive Learning Task}
%
Following~\citet{qiu2020gcc}, for each node type, we first sample sub-graphs from the heterogeneous graph via random walk starting from a start node in this node type.
Then, we regard the sub-graphs sampled from the same start node as positive pairs, and those from different start nodes as negative pairs.
Note that the start nodes of a negative pair should be of the same node type.
Finally, we apply the contrastive learning technique to train the GNN encoder by minimizing the InfoNCE loss~\citep{oord2018representation}:
\begin{equation}
\mathcal L_{\mathcal X} = - \log \frac{\exp({\mathbf{q}} ^\top \mathbf{k}_+  / \tau)}{\sum_{i=1}^K\exp({\mathbf{q}} ^\top \mathbf{k}_i  / \tau)}
\end{equation}
where $\mathcal X$ denotes a node set and $\mathcal X \in \{\mathcal T, \mathcal E, \mathcal W\}$,  $\mathbf{q}$ and $\mathbf{k}_+$ are the representations of two sub-graphs sampled from the same node $x^q \in \mathcal X$, and $\{\mathbf{k_0,k_1,\dots,k_K}\}$ is representations collection of other nodes except node $x^q$, 
which are generated by the GNN encoder $f$, denoted by $\mathbf{q}=f(x^q)$ and $\mathbf{k}=f(x^k)$.
In this way, the GNN encoder is guided to enhance the similarities (and dissimilarities) between positive (and negative) instances.

\vpara{Edge-level Link Reconstruction Learning Task}
%
Another task focuses on the relationships among different node types.
%
The motivation is that the representations of semantically similar textual content and emojis should also have higher similarity, even though text and emojis might come from the different distributions.
Based on the graph encoder $f$, we additionally use a score predictor that outputs the inner product of two nodes' representations as their similarity. If the similarity value exceeds a certain threshold, a edge between the text and emoji nodes is predicted to exist.

To construct training samples, we first sample some existing edges in the heterogeneous graph as positive examples and add an equal number of randomly sampled non-existing edges as negative examples. During training, the model is optimized using a cross-entropy loss to learn representations that maximize the similarity for positive examples and minimize the similarity for negative examples. The binary cross entropy loss for positive edges and negative sampled edges is computed as:
\begin{equation}
\mathcal{L}_{+}= -\frac{1}{|\mathcal{P}|} \sum_{(i,j) \in \mathcal{P}} \log p_{\theta}(y_{i,j}=1 | z_i, z_j) 
\end{equation}
\begin{equation}
\mathcal{L}_{-}= -\frac{1}{|\mathcal{N}|} \sum_{(i,j) \in \mathcal{N}} \log {(1 - p_{\theta}(y_{i,j}=1 | z_i, z_j))} 
\end{equation}
\begin{equation}
\mathcal{L} = \mathcal{L}_{+} + \mathcal{L}_{-}
\end{equation}
Where $\mathcal{P}$ is the set of positive edges, $\mathcal{N}$ is the set of negative edges, and $p_{\theta}(y_{i,j}=1 | z_i, z_j)$ is the probability that the edge $(i,j)$ exists.

For each edge type, we make the training processes separately, resulting in three sub-tasks, \emph{i.e.}, emoji reconstruction, important word identification and emoji-word position reconstruction.




\subsection{Downstream Tasks}
\label{subsec:downstreamtask}

To evaluate our proposed framework, we apply our emoji embeddings to two downstream tasks: popularity prediction and sentiment analysis.
For all downstream tasks, the task input is a social-media post $P = \{T_p; E_p\}$, where $T_p=\{w_{p_1}, w_{p_2}, ... w_{p_m}\}$ is the text in the post, $w_{p_i}$ means the $i$-th word in the post text $T_p$, and $m$ indicates the number of words in $T_p$. $E_p=\{e_{p_1}, e_{p_2}, ..., e_{p_n}\}$ denotes the emojis in $P$, $e_j$ is the $j$-th emoji, and $n$ represents the number of emojis in $E_p$.
%
The representations of each word $w_{p_i}$ and emoji $e_{p_j}$ can be initialized to $h_{w_{pi}} \in \mathbb{R}^d$ and $h_{e_{pj}} \in \mathbb{R}^d$ via our graph pre-training framework.\footnote{If a word or an emoji (in a downstream post) does not appear during pre-training, it will be discarded to avoid the out-of-vocabulary problem.} We first sample a sub-graph of this node from the heterogeneous graph via a random walk starting from the node, and then input the sub-graph into our GNN encoder model to get its representation, which also serves as the embedding of the node. This embedding is then used for downstream tasks.



\vpara{Popularity Prediction \& Sentiment Analysis}
These two tasks are both multi-class classification tasks, which take $P$ as input and output the popularity/sentiment prediction (assuming there are c classes). The process can be formulated as:
\begin{equation}
H_{e_p} = \sum^n_{i=1}h_{e_{pi}}
\end{equation}
\begin{equation}
H_{e_o} = h_t E^{T}E
\end{equation}
\begin{equation}
y_p = W_{pop} (h_t \oplus H_{e_p} \oplus H_{e_o})^{T}
\end{equation}
where $E\in \mathbb{R}^{K\times d}$ is the matrix embedding of the emoji set $\mathcal{E}$ (includes pre-trained embeddings of all emojis). $h_t \in \mathbb{R}^d$ is the embedding of the post text $T_p$, which is obtained from pre-trained language model backbones. $W_{pop} \in \mathbb{R}^{c\times 3d}$ is the trainable parameters. $y_p \in \mathbb{R}^c$ denote the predicted probability over all classes.

For objective, we choose the multi-class cross entropy, and we use the weighted labels to mitigate the imbalance of labels. The weight of labels is the inverse of their count. 
\begin{equation}
L_{pop} = - \sum_{i=1}^{c} w_{i} \hat{y}_{i} \log(y_{p_i}) \label{eq:weight}
\end{equation}
where $w_i$ is the label weight for the $i$-th category, $\hat{y}_i$ is the target and $y_{p_i}$ is the predicted probabilities for the $i$-th category.

\section{Experiment}
\label{sec:experiment}
In this section, we first introduce the implementation details of our proposed pre-training framework (\S~\ref{sec:pretrain}), and then evaluate our framework on two downstream tasks, \emph{i.e.}, popularity prediction (\S~\ref{sec:downstream task 1}) and sentiment classification (\S~\ref{sec:downstream task 2}). 


\begin{table*}[h]
    \centering
    \resizebox{0.7\textwidth}{!}{
    \begin{tabular}{lccccccc}
        \toprule
        Metric          & Mean  & Std    & Min   & .25   & .50   & .75   & Max     \\
        \midrule
        Liked Count     & 799.0 & 5198.5 & 0.0   & 66.0  & 148.0 & 334.0 & 357163.0 \\
        Collected Count & 468.3 & 3431.5 & 0.0   & 25.0  & 75.0  & 189.0 & 281984.0 \\
        Comments Count  & 46.3  & 140.7  & 0.0   & 9.0   & 22.0  & 46.0  & 8652.0   \\
        Emoji Count/Post& 9.3   & 8.4    & 2.0   & 4.0   & 7.0   & 12.0  & 316.0    \\
        Post Length     & 391.0 & 258.2  & 13.0  & 191.0 & 335.0 & 535.0 & 18500.0  \\
        \bottomrule
    \end{tabular}
    }
    \caption{Dataset Description Statistic.}
    \label{tab:dataset}
\end{table*}

\begin{table*}[h]
  \centering
  \huge
  \resizebox{0.90\textwidth}{!}
    {
  \begin{tabular*}{2\linewidth}{l|@{\extracolsep{\fill}}ccccc}
    \toprule
    \textbf{Backbone} & \textbf{BERT} & \textbf{GPT2} & \textbf{RoBERTa} & \textbf{OPT-1.3B} & \textbf{LlAMA2-7B} \\
    \midrule
    & \multicolumn{5}{c}{\emph{F1 score (Weighted) - Prediction of \# of likes}}\\
    \midrule
    Remove  & 0.3801 & 0.3203 & 0.3453 & 0.3602 & 0.4452 \\
    Keep V1    & 0.3697 & 0.3659 & 0.3702 & 0.3470 & 0.4489 \\
    Keep V2   & 0.3793 & 0.3535 & 0.3867 & 0.3578 & 0.4524 \\
    Translate       & 0.3820 & 0.3664 & 0.3442 & 0.3440 & 0.4362 \\
    Emoji2vec & 0.4357 & 0.3279 & 0.3421 & 0.3771 & 0.4878 \\
    Emoji Co-occurence & 0.3153 & 0.4009 & 0.4136 & 0.3999 & 0.4912 \\
    \textbf{Our model}   & \textbf{0.4487} & \textbf{0.4084} & \textbf{0.4444} & \textbf{0.4174} & \textbf{0.5120} \\
    \midrule
    & \multicolumn{5}{c}{\emph{F1 score (Weighted) - Prediction of \# of collects}}\\
    \midrule
    Remove  & 0.4060 & 0.3930 & 0.3924 & 0.4073 & 0.4132 \\
    Keep V1    & 0.3989 & 0.3887 & 0.3979 & 0.3890 & 0.4534 \\
    Keep V2   & 0.4003 & 0.3840 & 0.3966 & 0.3910 & 0.4487 \\
    Translate       & 0.3973 & 0.3929 & 0.3865 & 0.3881 & 0.4134 \\
    Emoji2vec   & 0.4377 & 0.3949 & 0.3927 & 0.4082 & 0.4695 \\
    Emoji Co-occurence & 0.3419 & 0.4307 & 0.4374 & 0.4414 & 0.4487 \\
    \textbf{Our model}   & \textbf{0.4387} & \textbf{0.4552} & \textbf{0.4680} & \textbf{0.4580} & \textbf{0.4776} \\
    \midrule
    & \multicolumn{5}{c}{\emph{F1 score (Weighted) - Prediction of \# of comments}}\\
    \midrule
    Remove  & 0.3816 & 0.3861 & 0.3854 & 0.3987 & 0.4134 \\
    Keep V1    & 0.3689 & 0.3556 & 0.3712 & 0.3806 & 0.4383 \\
    Keep V2   & 0.3863 & 0.3872 & 0.3769 & 0.3842 & 0.4313 \\
    Translate       & 0.3662 & 0.3805 & 0.3803 & 0.3881 & 0.4134 \\
    Emoji2vec   & 0.4317 & 0.3789 & 0.3893 & 0.3969 & 0.4631 \\
    Emoji Co-occurence & 0.2968 & 0.3891 & 0.4005 & 0.3894 & 0.4407 \\
    \textbf{Our model}   & \textbf{0.4353} & \textbf{0.4077} & \textbf{0.4256} & \textbf{0.4066} & \textbf{0.4695} \\
    \bottomrule
  \end{tabular*}
  }
  \caption{Experimental results on popularity prediction.}
  \label{tab:result}
\end{table*}

\subsection{Experimental setup}
\label{sec:pretrain}
During pre-training, we use Adam for optimization with a learning rate of 0.005, $\beta_1 = 0.9$, $\beta_2 = 0.999$, $\epsilon = 1\times10^8$, weight decay of $1e-4$, learning rate warm-up over the first 7,500 steps, and linear decay of the learning rate after 7,500 steps. Gradient norm clipping is applied with range $[-1,1]$. For MoCo~\cite{he2020momentum}, we use a mini-batch size of 32, dictionary size of 16,384, and momentum m of 0.999, the temperature $\tau$ is set as 0.07, and we adopt GIN~\cite{xu2018powerful} with 2 layers and 768 hidden units each layer as our encoders. 
The GNN encoder trained in two types of pre-training tasks is then used to generate embeddings of post, emoji and text for downstream tasks.

\subsection{Popularity Prediction}
\label{sec:downstream task 1}

\vpara{Dataset}
Our first dataset mainly comes from the Chinese social media platform Xiaohongshu\footnote{\url{https://www.xiaohongshu.com/}}. We collect over 4 million Xiaohongshu posts through web crawling, and all the information are desensitized. In addition to the content of the post, other information mainly includes the author’s Xiaohongshu ID, post title (text information), number of likes, number of collects, number of comments, and publishing time. The user information mainly includes the number of followers, the total number of likes, and whether the account is an official certified account.
Note that this Xiaohongshu dataset will not be public due to ethical considerations (\emph{e.g.}, the post might contain individuals' privacy).
Our popularity label is derived from the ``liked count'', ``collected count'', and ``comments count'' fields in the Xiaohongshu dataset. Given the continuous nature of this data, to simplify it into a categorical task, we cluster them based on the distribution trend of popularity according to the data percentile of [0-50), [50-80), and [80-100], resulting in three categories (low, medium, and high). Thus, the ratio of category labels is low:medium:high = 5:3:2, indicating the imbalanced labels in the data. Given this, we use the weighted labels to mitigate the imbalance issue. The weights of the labels are the inverses of their counts, as described in Eq.~\ref{eq:weight}. The statistic of dataset is shown in table~\ref{tab:dataset}.


\vpara{Compared Baselines}
To evaluate the performance of our pre-training framework, we also employ two types of emoji processing methods as baselines:

\begin{itemize}[leftmargin=*,topsep=0pt]
\setlength{\itemsep}{0pt}
\setlength{\parsep}{0pt}
\setlength{\parskip}{0pt}
\item \textbf{Preprocessing methods:} We use the traditional emoji preprocessing methods introduced in section~\ref{sec:related} as baselines: 
(1) \emph{Remove} emojis: we use vectors of zeroes as the embedding of emojis to eliminate the information contained in them. 
(2) \emph{Keep} emojis: we use vectors of numbers randomly generated from Gaussian distribution as the embedding of emojis, meaning that they are different tokens but the model has no prior information on emojis. We generated two versions of embeddings to avoid randomness, which are denoted as \emph{Keep V1} and \emph{Keep V2}.
(3) \emph{Translate} emojis: we translate emojis into their descriptions (\emph{e.g.}, replace ``\emojicar'' with the word ``car''), and then use our backbone LLMs to encode emojis. In this way, emojis are treated just as normal words.

\item \textbf{Emoji representation learning method:}
These baseline model utilize the \emph{Emoji2vec}~\citep{Eisner2016emoji2vecLE} and the \emph{Emoji Co-occurence}~\citep{emojicoocurrence} methods to obtain emojis' embedding. For those emojis not included in the model, we initiate them to vectors of zeroes.
\end{itemize}

In all popularity prediction methods, the embedding of text in the post $T_p$ is encoded by corresponding different backbones (\emph{i.e.}, BERT~\cite{devlin2018bert}, GPT2~\cite{radford2019language}, RoBERTa~\cite{liu2019roberta}, OPT-1.3B~\cite{zhang2022opt} and Llama2-7B~\cite{touvron2023llama}). With regard to baseline models, we do not use DeepMoji~\cite{deepmoji,park-etal-2018-plusemo2vec} and  SEntiMoji~\cite{10.1145/3338906.3338977} for they simply use emojis as annotations and do not learn the representation of emojis. Furthermore, the majority of emojis are represented by Unicode characters that LLMs cannot recognize. Therefore, the translate method mentioned above is essentially an enhanced approach of directly inputting the original text containing emojis into an LLM. Consequently, we did not include the latter in the baseline.

\vpara{Results \& Analyses}
As the dataset is quite imbalanced on each label, we use the \emph{weighted F1 score} as an evaluation metric. 
To get a robust result, we run each task 10 times with different random initiations and summarize its average metrics.

As illustrated in Table~\ref{tab:result}, our method outperforms all four baselines on all three sub-types of popularity prediction tasks. This finding suggests that emojis offer extra information for popularity prediction by comparing our model with the \emph{remove} baseline. Besides, traditional methods generally fail or have a minor effect on extracting information from emojis.
When the emoji embedding is initiated by Gaussian random noise, the F1 score even decreases. This is because no improvement in the prediction performance can be attributed to the absence of semantic information in Gaussian noise.
The \emph{translate} baseline utilizes LLMs to extract emoji information by translating emojis into words and then encoding them. Different from Gaussian noise, emoji tokens contain valuable semantics.
However, due to there is only a very small part of emojis can be translated, we can infer that incorporating emoji information results in nearly no improvement in the F1 score.
\emph{Emoji2vec} enhances the extraction of emoji information and achieves better performance compared to \emph{translate}.
Our method shows the highest F1 score among all models, indicating its strong capacity for extracting information from emojis. Detailed precision and recall results are shown in appendix~\ref{sec:appendix1}. 

To further demonstrate the superior performance of our model over baseline models on downstream tasks, we conducted two-sample t-tests between our model and each baseline model. The result shows that our model significantly outperforms baseline models at the 0.05 significance level. Detailed results of the hypothesis tests are shown in Table~\ref{tab:t-test}.

\begin{figure}[h]
  \centering  
  \includegraphics[width=1\linewidth]{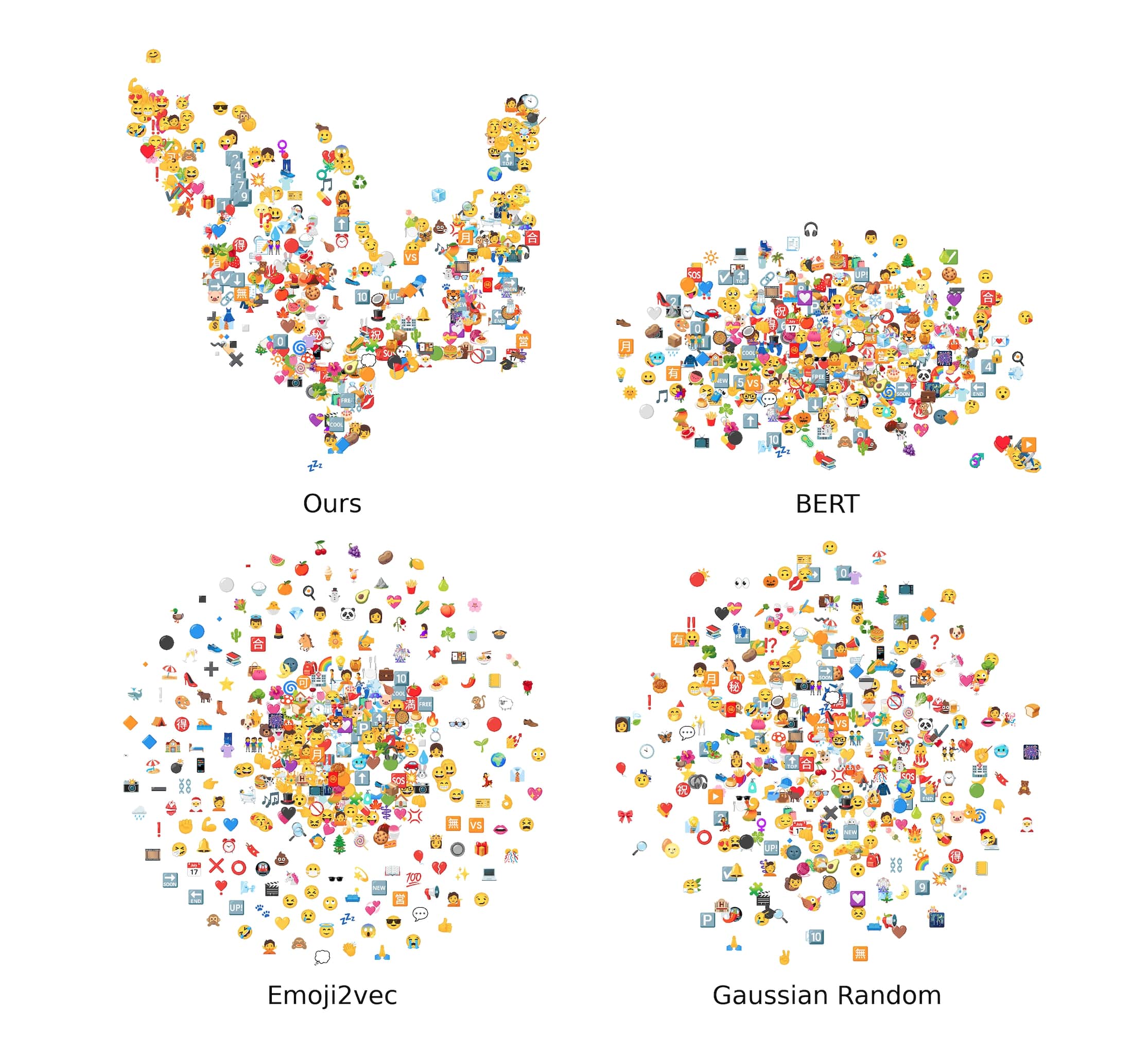}  
  \caption{The t-SNE visualization results of emoji embeddings from four models.}
  \label{fig:tsne}
\end{figure}

To intuitively observe the difference among these methods for extraction of the emoji embedding, we conduct the t-SNE~\cite{JMLR:v9:vandermaaten08a} visualization for the generated emoji embeddings.
As shown in Figure~\ref{fig:tsne}, the embeddings extracted by other models are randomly distributed in the 2D embedding space, whereas our method effectively groups similar emojis together. For example, emojis representing happiness are situated in the upper left area, emojis denoting sad emotions are located in the upper right area, number emojis are clustered together, and object emojis populate the lower region.

\begin{table*}[h]
  \centering
  \large
  \resizebox{0.83\textwidth}{!}
    {
  \begin{tabular*}{\linewidth}{l|@{\extracolsep{\fill}}ccccc}
    \toprule
    \textbf{Backbone} & \textbf{BERT} & \textbf{GPT2} & \textbf{RoBERTa} & \textbf{OPT-1.3B} & \textbf{LlAMA2-7B} \\
        \midrule
        & \multicolumn{5}{c}{\emph{F1 score (Weighted) - Prediction of \# of likes}}\\
        \midrule
        v.s. Remove    & 0.0013 & 0.0000 & 0.0000 & 0.0000 & 0.0000 \\
        v.s. Keep V1   & 0.0344 & 0.0023 & 0.0000 & 0.0000 & 0.0000 \\
        v.s. Keep V2   & 0.0000 & 0.0015 & 0.0000 & 0.0002 & 0.0000 \\
        v.s. Translate& 0.0000 & 0.0003 & 0.0000 & 0.0001 & 0.0000 \\
        v.s. Emoji2vec   & 0.0462 & 0.0000 & 0.0000 & 0.0000 & 0.0000 \\
        \midrule
        & \multicolumn{5}{c}{\emph{F1 score (Weighted) - Prediction of \# of comments}}\\
        \midrule
        v.s. Remove    & 0.0416 & 0.0000 & 0.0000 & 0.0000 & 0.0014 \\
        v.s. Keep V1   & 0.0000 & 0.0000 & 0.0000 & 0.0000 & 0.0000 \\
        v.s. Keep V2   & 0.0000 & 0.0000 & 0.0000 & 0.0000 & 0.0000 \\
        v.s. Translate& 0.0000 & 0.0000 & 0.0000 & 0.0000 & 0.0000 \\
        v.s. Emoji2vec   & 0.0693 & 0.0000 & 0.0000 & 0.0000 & 0.0014 \\
        \midrule
        & \multicolumn{5}{c}{\emph{F1 score (Weighted) - Prediction of \# of collects}}\\
        \midrule
        v.s. Remove    & 0.0257 & 0.0000 & 0.0000 & 0.0087 & 0.0052 \\
        v.s. Keep V1   & 0.0000 & 0.0000 & 0.0000 & 0.0000 & 0.0000 \\
        v.s. Keep V2   & 0.0000 & 0.0000 & 0.0000 & 0.0000 & 0.0000 \\
        v.s. Translate& 0.0000 & 0.0000 & 0.0000 & 0.0000 & 0.0000 \\
        v.s. Emoji2vec   & 0.0482 & 0.0000 & 0.0000 & 0.0329 & 0.0020 \\
        \bottomrule
      \end{tabular*}
  }
  \caption{Results of t-test. Each score represents the average p-value for t-test.}
  \label{tab:t-test}
\end{table*}

\begin{table}[t]
\centering
\resizebox{0.45\textwidth}{!}{
\begin{tabular}{p{3.5cm}>{\centering\arraybackslash}p{3.5cm}}
\toprule
Model & F1 score (weighted)\\
\midrule
Remove & 0.5446\\
Keep V1 & 0.4336\\
Keep V2 & 0.4521\\
Translate & 0.4758\\
Emoji2vec & 0.5507\\
Emoji Co-occurence & 0.5297\\
SEntiMoji & 0.5512\\
DeepMoji & 0.5294 \\
\textbf{Our model} & \textbf{0.5679}\\
\bottomrule
\end{tabular}
}
\caption{Experimental results on sentiment classification.}
\label{tab:result2}
\end{table}

\begin{table*}[h]
  \centering
  \resizebox{0.90\textwidth}{!}
    {
  \begin{tabular*}{\linewidth}{lcc}
    \toprule
     \textbf{Text} & \textbf{Prediction} & \textbf{Ground Truth}  \\
    \midrule
         Maye got me remembering that shit. Talking about scams. \emojiptwx    & Negative & Positive
    \\
         I'm kinda spoiled.  \emojiseenoevil\emojiconfused\emojicrown  & Positive & Negative
    \\
     I miss you too. \emojisob\emojidancers\emojiheart\emojiwatermelon  & Positive & Negative
     \\
    \bottomrule
      \end{tabular*}
  }
  \caption{Some bad cases in sentiments analysis.}
  \label{tab:badcases}
\end{table*}

\begin{table*}[h]
\centering
\resizebox{0.6\textwidth}{!}{
\begin{tabular}{lc}
\toprule
Model & F1 score (weighted) \\
\midrule
Original framework & \textbf{0.4691} \\
- Node level contrastive learning: Post & 0.4529 \\
- Node level contrastive learning: Word & 0.4497 \\
- Node level contrastive learning: Emoji & 0.4434 \\
- Edge level link reconstruction: Post - Emoji & 0.446 \\
- Edge level link reconstruction: Word - Post & 0.4507 \\
- Edge level link reconstruction: Emoji - Word & 0.4401 \\
\bottomrule
\end{tabular}
}
\caption{Ablation study on node level and edge level tasks. Full model outperforms all other models, meaning that every part of our model is useful.}
\label{tab:ablation}
\end{table*}

\vpara{Ablation Study}
The node-level sub-graph contrastive learning task is separately conducted on different types of nodes, and thus includes three sub-tasks, \emph{i.e.}, post-node, word-node and emoji-node contrastive learning. The edge-level link reconstruction also consists of three sub-tasks, \emph{i.e.}, post-emoji, word-post and emoji-word link reconstruction.

To evaluate the effectiveness of these sub-tasks, we conduct ablations by removing each of them in our pre-training framework, and evaluate the model performance on popularity prediction (number of likes).
%
As shown in Table~\ref{tab:ablation}, the performance of our original framework exceeds every other variant model, indicating that our model gain information from every part of pre-training tasks.
In conclusion, the ablation study shows that Emoji representation learning can benefit from all the pre-training sub-tasks, leading to enhanced prediction performance.

\subsection{Sentiment Classification}
\label{sec:downstream task 2}

\vpara{Dataset} The dataset used for the sentiment classification task, is a public dataset consisting of Twitter posts generated by authors of Emoji2vec~\citep{Eisner2016emoji2vecLE}, which is also used in their experiments on Emoji2vec. This dataset inherently includes three types of sentiment labels for each post: negative, neutral, and positive. During data preprocessing, we filtered out tweets without emojis from the dataset, separated emojis and text for the rest, and then used BERT to obtain post embeddings.

\vpara{Compared Baselines}
To evaluate the performance of our pre-training framework, we employ the same baselines as those in section~\ref{sec:downstream task 1}. Additionally, we also add SEntiMoji and DeepMoji in our baseline models. Although they do not generate emoji embeddings which can be applied in various downstream tasks, they still serve as strong baselines in sentiment classification task.

\vpara{Results \& Analyses}
As shown in Table~\ref{tab:result2}, our method outperforms all baselines on the sentiment classification task with three categories, achieving a F1 score of 0.5679.
%
SEntiMoji and Emoji2vec performs well, while other baseline models fail to extract emoji information in the posts.

\vpara{Bad Cases Study}
In this bad case study of sentiment analysis task, which are shown in Table~\ref{tab:badcases}, we observe discrepancies between the model's predictions and the actual sentiments. These instances highlight several critical challenges.

\begin{itemize}[leftmargin=*,topsep=0pt]
\setlength{\itemsep}{0pt}
\setlength{\parsep}{0pt}
\setlength{\parskip}{0pt}

\item \textbf{Understanding context and culture:}  
In the first case, where our method predicted the sentiment as Negative while the ground truth was Positive, the instance likely represents a form of sarcasm or humor. However, in a Chinese context, the emoji \emojiptwx may also carry the connotation of tears of helplessness, pointing towards a complexity involving both the understanding of context and the diverse meanings emojis can have across different linguistic cultures.

\item \textbf{Ambiguity of emojis and complexity of emotions:}
In both the second and third challenging scenarios, the sentences feature a mix of emojis, each imbued with either positive or negative emotional undertones. This blend introduces an intricate layer to the overall sentiment conveyed, making the emotional semantics of the sentence more complex. Although humans may find it relatively straightforward to analyze and understand the sentiment of such sentences, models might struggle to make accurate determinations in these nuanced situations.

\end{itemize}

\section{Conclusion}
\label{sec:conclusion}
In this paper, we propose a novel framework for joint pre-training on text and emojis. To model the relationship between posts, words and emojis, we first construct a heterogeneous graph comprising three node types corresponding to each element. Subsequently, we propose a subgraph-level pre-training framework for representation learning, including node-level graph contrastive learning and edge-level link reconstruction learning.
Our experimental results on the two downstream tasks, \emph{i.e.}, popularity prediction and sentiment analysis, demonstrate that these pre-training tasks could leverage the acquired embeddings for downstream applications, and our approach yields substantial improvements over baseline methods.

\section*{Limitations}

We discuss the limitations and how we can potentially address them in this section.

\noindent \textbf{Limitations in Emoji Modalities:} While our proposed model could handle the majority of emoji representations, including Unicode encoding and text-based descriptions, it currently lacks the capability to process emoji in image format. This limitation is noteworthy, especially considering the prevalent use of image-based emojis on certain social media platforms. Our future endeavors involve exploring multimodal approaches to enhance our model's versatility.

\noindent \textbf{Data Cleansing Requirements:} The efficacy of our model hinges on accurate emoji tokenization in data preprocessing. Under certain extreme conditions, such as processing an extensive volume of emojis that resist correct tokenization, the model's performance may be compromised. Addressing these challenges remains a focus for future improvements.

\section*{Ethics Statement}

There are no ethics-related issues in this paper. The Xiaohongshu data is used only for scientific research purposes, and will not be public. The Twitter data and other related resources in this work are open-source and commonly-used by many existing work.
\section*{Acknowledgements}

This work was supported in part by NSFC (62206056, 92270121). Additionally, we are grateful to  Yang Yang for providing the necessary resources that made this research possible. Finally, we would like to thank the anonymous reviewers for their thoughtful comments and suggestions that helped improve the quality of this paper.

\bibliography{reference}

\appendix



\section{Full Results of Popularity Prediction}
\label{sec:appendix1}

Table~\ref{tab:recall} and Table~\ref{tab:precision} show the recall and precision of our model and baseline models, which also verify the superiority of our model.

\begin{table*}[h]
  \centering
  \resizebox{0.90\textwidth}{!}
    {
  \begin{tabular*}{\linewidth}{l|@{\extracolsep{\fill}}ccccc}
    \toprule
    \textbf{Backbone} & \textbf{BERT} & \textbf{GPT2} & \textbf{RoBERTa} & \textbf{OPT-1.3B} & \textbf{LlAMA2-7B} \\
    \midrule
    & \multicolumn{5}{c}{\emph{Recall (Weighted) - Prediction of \# of likes}}\\
    \midrule
    Keep V1 & 0.4056 & 0.3912 & 0.4017  & 0.3649 & 0.4650 \\
    Keep V2 & 0.4180 & 0.3766 & 0.4251  & 0.3698 & 0.4634 \\
    Remove  & 0.4375 & 0.3879 & 0.3703  & 0.3875 & 0.4952 \\
    Translate & 0.4060 & 0.4343 & 0.4237  & 0.3807 & 0.4588 \\
    Emoji2vec & 0.4338 & 0.3871 & 0.4044  & 0.3788 & 0.4884 \\
    Emoji Co-occurence & 0.3970 & 0.4047 & 0.4253 & 0.4183 & 0.5001 \\
    \textbf{Our model}   & \textbf{0.4464} & \textbf{0.4109} & \textbf{0.4480} & \textbf{0.4190} & \textbf{0.5150} \\
    \midrule
    & \multicolumn{5}{c}{\emph{Recall (Weighted) - Prediction of \# of collects}}\\
    \midrule
    Keep V1 & 0.4259 & 0.4158 & 0.4145  & 0.4223 & 0.4588 \\
    Keep V2 & 0.4101 & 0.4034 & 0.4127  & 0.4077 & 0.4572 \\
    Remove  & 0.4314 & 0.3949 & 0.3952  & 0.4019 & 0.4610 \\
    Translate & 0.4092 & 0.4196 & 0.4046  & 0.4031 & 0.4653 \\
    Emoji2vec & 0.4338 & 0.3993 & 0.3973  & 0.4043 & 0.4710 \\
    Emoji Co-occurence & 0.4009 & 0.4424 & 0.4471 & 0.4452 & 0.4648 \\
    \textbf{Our model}   & \textbf{0.4342} & \textbf{0.4566} & \textbf{0.4656} & \textbf{0.4556} & \textbf{0.4763} \\
    \midrule
    & \multicolumn{5}{c}{\emph{Recall (Weighted) - Prediction of \# of comments}}\\
    \midrule
    Keep V1 & 0.4003 & 0.3976 & 0.3860  & 0.3989 & 0.4454 \\
    Keep V2 & 0.4087 & 0.4110 & 0.4111  & 0.4008 & 0.4422 \\
    Remove  & 0.4282 & 0.3987 & 0.3970  & 0.3933 & 0.4625 \\
    Translate & 0.3935 & 0.4164 & 0.3975  & 0.4031 & 0.4653 \\
    Emoji2vec & 0.4319 & 0.4021 & 0.3989  & 0.3975 & 0.4621 \\
    Emoji Co-occurence & 0.3810 & 0.4137 & 0.4120 & 0.4087 & 0.4531 \\
    \textbf{Our model}   & \textbf{0.4344} & \textbf{0.4212} & \textbf{0.4322} & \textbf{0.4126} & \textbf{0.4717} \\
    \bottomrule
  \end{tabular*}
  }
  \caption{Recall of experiments on popularity prediction.}
  \label{tab:recall}
\end{table*}

\begin{table*}[h]
  \centering
  \resizebox{0.90\textwidth}{!}
    {
  \begin{tabular*}{\linewidth}{l|@{\extracolsep{\fill}}ccccc}
    \toprule
    \textbf{Backbone} & \textbf{BERT} & \textbf{GPT2} & \textbf{RoBERTa} & \textbf{OPT-1.3B} & \textbf{LlAMA2-7B} \\
    \midrule
    & \multicolumn{5}{c}{\emph{Precision (Weighted) - Prediction of \# of likes}}\\
    \midrule
    Keep V1 & 0.3947 & 0.3948 & 0.3805  & 0.3979 & 0.4710 \\
    Keep V2 & 0.4160 & 0.3933 & 0.3790  & 0.3963 & 0.4754 \\
    Remove  & 0.4766 & 0.3872 & 0.4246  & 0.4421 & 0.5180 \\
    Translate & 0.4137 & 0.3963 & 0.3435  & 0.3957 & 0.4442 \\
    Emoji2vec & 0.4755 & 0.3923 & 0.3659  & 0.4412 & 0.5103 \\
    Emoji Co-occurence & 0.2839 & 0.4117 & 0.4270 & 0.4139 & 0.5008 \\
    \textbf{Our model}   & \textbf{0.4808} & \textbf{0.4248} & \textbf{0.4603} & \textbf{0.4350} & \textbf{0.5207} \\
    \midrule
    & \multicolumn{5}{c}{\emph{Precision (Weighted) - Prediction of \# of collects}}\\
    \midrule
    Keep V1 & 0.4130 & 0.3937 & 0.3965  & 0.3941 & 0.4621 \\
    Keep V2 & 0.4190 & 0.4147 & 0.3967  & 0.3999 & 0.4598 \\
    Remove  & 0.4518 & 0.4102 & 0.4256  & 0.4223 & 0.4715 \\
    Translate & 0.4161 & 0.3990 & 0.3948  & 0.4018 & 0.4172 \\
    Emoji2vec & 0.4531 & 0.4099 & 0.4236  & 0.4206 & 0.4779 \\
    Emoji Co-occurence & 0.3601 & 0.4333 & 0.4579 & 0.4506 & 0.4523 \\
    \textbf{Our model}   & \textbf{0.4525} & \textbf{0.4587} & \textbf{0.4804} & \textbf{0.4645} & \textbf{0.4842} \\
    \midrule
    & \multicolumn{5}{c}{\emph{Precision (Weighted) - Prediction of \# of comments}}\\
    \midrule
    Keep V1 & 0.3950 & 0.3648 & 0.3909  & 0.3989 & 0.4388 \\
    Keep V2 & 0.3945 & 0.4005 & 0.3746  & 0.4080 & 0.4343 \\
    Remove  & 0.4424 & 0.4003 & 0.4204  & 0.4107 & 0.4713 \\
    Translate & 0.4176 & 0.3977 & 0.4071  & 0.4018 & 0.4172 \\
    Emoji2vec & 0.4419 & 0.3965 & 0.4209  & 0.4075 & 0.4716 \\
    Emoji Co-occurence & 0.3045 & 0.3878 & 0.4008 & 0.3892 & 0.4441 \\
    \textbf{Our model}   & \textbf{0.4450} & \textbf{0.4043} & \textbf{0.4254} & \textbf{0.4061} & \textbf{0.4732} \\
    \bottomrule
  \end{tabular*}
  }
  \caption{Precision of experiments on popularity prediction.}
  \label{tab:precision}
\end{table*}

\end{document}